\documentclass[letterpaper, 10 pt, conference]{ieeeconf}  % Comment this line out if you need a4paper

\IEEEoverridecommandlockouts                              % This command is only needed if 
\usepackage{graphics} % for pdf, bitmapped graphics files
\usepackage{epsfig} % for postscript graphics files
\usepackage{times} % assumes new font selection scheme installed
\usepackage{amsmath} % assumes amsmath package installed
\usepackage{amssymb}  % assumes amsmath package installed

\usepackage{booktabs}
\usepackage{multirow}

\title{\LARGE \bf
CARLA-GS: Decoupling Representation, Reasoning, and Physics Simulation for Autonomous Driving Corner-Case Synthesis
}

\author{Kaicong Huang$^{1}$, Meng Ma$^{1}$, Ruimin Ke$^{1*}$% <-this % stops a space
\thanks{$^{1}$ Department of Civil and Environmental Engineering, Rensselaer Polytechnic Institute, 110 Eighth Street, Troy, NY USA 12180.}%
\thanks{$^{*}$ Corresponding author. Email: \tt\small ker@rpi.edu}%
}

\begin{document}

\maketitle
\thispagestyle{empty}
\pagestyle{empty}

\begin{abstract}
Safety evaluation for autonomous driving is dominated by rare, safety-critical interactions, motivating simulators that can deliberately synthesize corner cases with photorealistic observations. Corner-case generation is inherently a multi-source problem spanning visual representation, scene reasoning, and vehicle trajectory generation and control. Prior knowledge- and model-based approaches typically focus on scene or trajectory components in isolation, while diffusion-based methods attempt end-to-end generation but still struggle to ensure spatiotemporal consistency and physical realism.
To unify these aspects within a single framework, we propose CARLA-GS, a modular corner-case synthesis pipeline that decouples visual representation, semantic reasoning, and physics-based execution while maintaining tight cross-module coupling. Starting from real driving data, we reconstruct an editable gaussian scene with additional geometry-consistent constraints. A multi-agent LLM then performs scene-level reasoning to identify risky interactions and generate intent-level waypoint trajectories, while the low-level motion control is delegated to CARLA, where a PID controller ensures kinematic and dynamic feasibility. The simulated vehicle states are finally re-projected into the gaussian scene for ego-centric rendering. This design enables high-level semantic reasoning, low-level physically executable motion, and photorealistic corner-case generation within a unified pipeline.
Experiments on the Waymo Open Dataset show, both quantitatively and qualitatively, that our framework enables controllable corner-case generation and produces photorealistic, spatiotemporally consistent videos aligned with semantic intent and physically feasible motion. Code is available at https://github.com/ccoopq/CARLA-GS.
\end{abstract}

\begin{figure*}[t]
    \centering
    \includegraphics[width=1.0\textwidth]{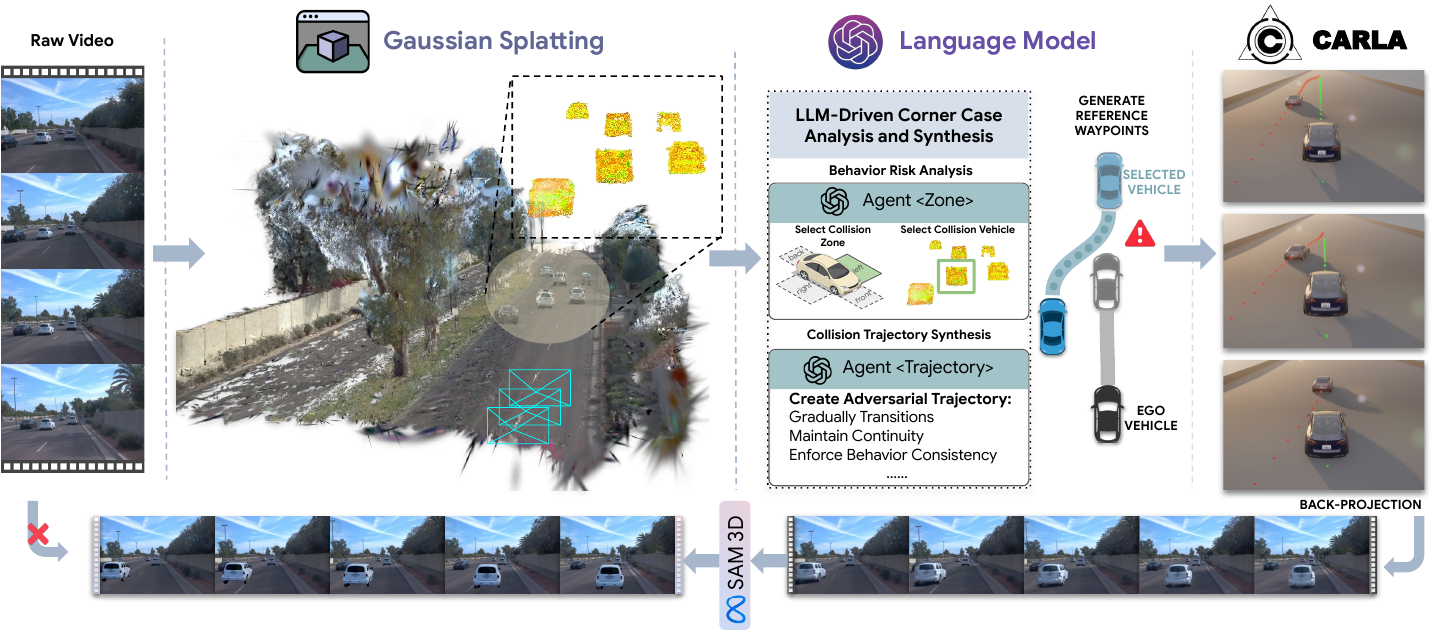}
    \caption{Overall framework of CARLA-GS. We decouple scene representation, risk reasoning, and physics simulation to leverage the strengths of each module. Specifically, 3D Gaussian Splatting provides high-fidelity scene reconstruction and rendering, while vehicle states are fed to an LLM with chained $<$Zone$>$ and $<$Trajectory$>$ agents to synthesize risk-aware scenarios. The LLM outputs only target trajectories, and low-level control is handled in CARLA via a PID controller. Finally, simulated vehicle poses are re-projected into the Gaussian scene to render geometrically consistent and physically realistic corner-case videos. By replacing raw objects with priors generated by 3D foundation models, rendering artifacts under novel viewpoints can be further mitigated.}
    \label{fig:framework}
\end{figure*}

\section{INTRODUCTION}
Autonomous driving (AD) operates in open-world traffic where safety evaluation is inherently a rare-event problem, known as the Curse of Rarity~\cite{liu2024curse}. This motivates simulation pipelines that intentionally expose corner cases. Meanwhile, camera-centric AD systems require simulators to provide not only trajectory rollouts but also photorealistic observations for reliable closed-loop evaluation~\cite{wang2025omnidrive}.

This requirement has renewed interest in scalable driving-scene reconstruction for safety-oriented simulation. NeRF-based methods achieve high-fidelity view synthesis but remain expensive for large-scale optimization and rendering~\cite{gao2022nerf}, while diffusion-based world models provide strong generative diversity yet still struggle with physical plausibility and multi-view consistency under real-time constraints~\cite{wang2024drivedreamer,russell2025gaia}. In contrast, 3D/4D Gaussian Splatting (GS)~\cite{kerbl20233d,wu20244d} enables real-time differentiable rendering with explicit scene primitives, making it attractive for urban driving simulation~\cite{zhou2024drivinggaussian,yan2024street,xu2025cruise}. However, vanilla 3DGS is designed for photometric reconstruction, lacking instance-level control over traffic participants and remaining sensitive to reconstruction artifacts that may affect occlusion or collision outcomes~\cite{huang20242d,chen2024pgsr}.

Moreover, corner-case generation is not purely a visual problem. It requires semantic reasoning over agent intent, interaction context, and future motion evolution. Prior work synthesizes corner cases through data-driven, knowledge-based, or learning-based pipelines~\cite{zhang2025drivegen,drayson2023cc,lu2024realistic}. However, these approaches often lack semantic scene understanding and high-level decision reasoning for identifying plausible risky interactions, limiting their flexibility in generating targeted behaviors. Inspired by~\cite{mei2025seeking}, we adopt LLM-based reasoning as a semantic planning module for corner-case synthesis. LLMs provide macro-level scene understanding and prompt-driven behavioral control, enabling intent-level risk specification while automatically selecting interacting agents from the scene.

After scene-level analysis, the next step is to generate vehicle trajectories and execute them under realistic kinematic and dynamic constraints. While LLMs are effective for high-level trajectory planning, they are not well suited to produce dynamically feasible control signals or guarantee collision-valid outcomes. Embedding physics directly into 3DGS is also challenging, as Gaussian primitives are unstructured and lack the rigid topology required by conventional dynamics solvers. Although recent works augment Gaussians with physical attributes~\cite{xie2024physgaussian,huang2026real2sim,jiang2024vr} or reconstruct simulation scenes via meshes~\cite{xie2025vid2sim,zhu2025vr}, faithful vehicle dynamics simulation remains limited. This motivates a modular design that delegates physics-consistent motion to dedicated simulators such as CARLA~\cite{dosovitskiy2017carla}.

To overcome the above limitations, we propose \textbf{CARLA-GS}, a modular pipeline for safety-critical corner-case synthesis. The key idea is to \textbf{decouple visual representation, semantic reasoning, and physics-based simulation} while preserving cross-module consistency.
Specifically, we adopt Street Gaussians~\cite{yan2024street} as the backbone, whose object-background modeling strategy provides a strong foundation for instance-level vehicle editing. We further introduce geometry-consistent constraints to improve structural coherence of the reconstructed scene. In parallel, a multi-agent LLM performs semantic scene analysis to identify the collision zone around the ego vehicle and select a target adversarial vehicle, and then generates an intent-level corner-case trajectory in waypoint form. The resulting trajectories are then executed in CARLA, where a PID controller enforces kinematic and dynamic feasibility. Finally, the simulated vehicle states are re-injected into the geometry-consistent 3DGS scene, refined with a 3D foundation model, to render ego-centric photorealistic observations. The overall framework is shown in Fig.~\ref{fig:framework}.
This design leverages the complementary strengths of each module: Gaussian Splatting provides spatiotemporal consistency and high-fidelity visual representation, LLMs enable high-level semantic reasoning and flexible control, and CARLA ensures reliable vehicle physics, resulting in a controllable and scalable framework for corner-case generation.

Our main contributions are summarized as follows:
\begin{itemize}
\item We present CARLA-GS, a modular framework for driving corner-case synthesis that decouples visual representation, semantic reasoning, and physics simulation, enabling spatiotemporally consistent video generation.

\item We design a multi-agent LLM reasoning module for semantic scene understanding and intent-level trajectory planning, enabling automatic discovery of risky interactions and flexible prompt-driven adversarial control.

\item We develop a plug-and-play CARLA-Gaussian integration through state re-projection, allowing physically simulated vehicle motions to be reflected in photorealistic renderings.
\end{itemize}

\section{RELATED WORK}

\subsection{Autonomous Driving Scene Reconstruction and Corner-Case Generation}

A practical autonomous driving simulator should provide realistic perception inputs and support spatio-temporal reasoning~\cite{fruhwirth2025stsbench}. Neural scene representations enable photorealistic rendering from captured data, where NeRF models scenes as implicit radiance fields~\cite{mildenhall2021nerf,gao2022nerf} and can be enhanced with structured depth or scene priors for improved novel-view synthesis~\cite{li2021mine,wu2023mapnerf,li2023lanesegnet}. Recently, diffusion-based world models have also been explored for scalable and controllable driving-scene generation~\cite{wang2024drivedreamer,russell2025gaia}. However, NeRF-based pipelines are often computationally expensive for large-scale simulation, while diffusion models may hallucinate structures and struggle to maintain strict geometric consistency and deterministic control.

More recently, 3D Gaussian Splatting (3DGS) has emerged as an efficient alternative by representing scenes with explicit Gaussian primitives and enabling high-throughput rendering~\cite{zhu20243d}. It has been extended to driving-scene synthesis, editing, dynamic modeling, real-time SLAM, and model alignment~\cite{zhou2024drivinggaussian,peng2025desire,xu2025cruise,kulhanek2025lodge,wu20244d,bai2025rp}. Nevertheless, for safety-oriented corner-case synthesis, vanilla 3DGS remains insufficient because it lacks guarantees on geometric stability, actionable semantics, and physically consistent interactions~\cite{huang20242d,chen2024pgsr}.

Corner cases expose safety-critical behaviors that rarely appear in naturalistic driving data. Existing generation methods can be broadly divided into data-driven, knowledge-based, and LLM-based approaches. Data-driven methods synthesize realistic safety-critical scenarios from large-scale driving datasets~\cite{wang2024deepaccident,wang2025terasim,zhang2025drivegen}, but often lack explicit control over failure modes and interaction intent. Knowledge-based methods transform common scenarios into corner cases using structured representations such as scene graphs~\cite{drayson2023cc}, while their diversity and realism are limited by the coverage and correctness of symbolic rules. Recent LLM-based methods introduce high-level reasoning and multimodal understanding for generating risky scenarios or adversarial behaviors~\cite{lu2024realistic,mei2025seeking}. However, translating semantic intent into dynamically feasible vehicle behaviors remains challenging. Our work addresses these limitations by grounding LLM-driven reasoning in CARLA's physics-executable backend and integrating 3DGS-based visual reconstruction into a modular yet integrated pipeline for controllable and scalable corner-case generation.

\section{METHODOLOGY}

\subsection{Geometry-Consistent Driving Scenario Reconstruction}
\label{subsec:geo}
Due to the limited viewpoints of onboard driving cameras (typically 3-6 poses), vanilla 3DGS often produces geometrically inconsistent and bumpy surfaces~\cite{xie2025vid2sim, sun2020scalability}. We therefore introduce geometry-consistent regularization to enforce surface continuity and smoothness.

\textbf{Flattening Constraints.}
Following prior 2DGS and surface reconstruction works~\cite{huang20242d, chen2024pgsr, zhu2025vr}, we introduce a flattening regularization by constraining the anisotropic scaling of each Gaussian. The covariance of the $i$-th Gaussian is parameterized by a diagonal scaling matrix $\mathbf{S}_i=\mathrm{diag}(s_{i1}, s_{i2}, s_{i3})$, where $s_{i1}, s_{i2}, s_{i3}>0$ denote the principal-axis scales and $N$ is the total number of Gaussians. The flattening loss is defined as
\begin{equation}
\mathcal{L}_{\text{flat}} =
\frac{1}{N}\sum_{i=1}^{N}
\min\left(s_{i1},\, s_{i2},\, s_{i3}\right)
\end{equation}
which encourages each Gaussian to collapse along its least significant axis and thus promotes locally smooth surface structures.

\begin{figure}[t]
    \centering
    \includegraphics[width=1.0\columnwidth]{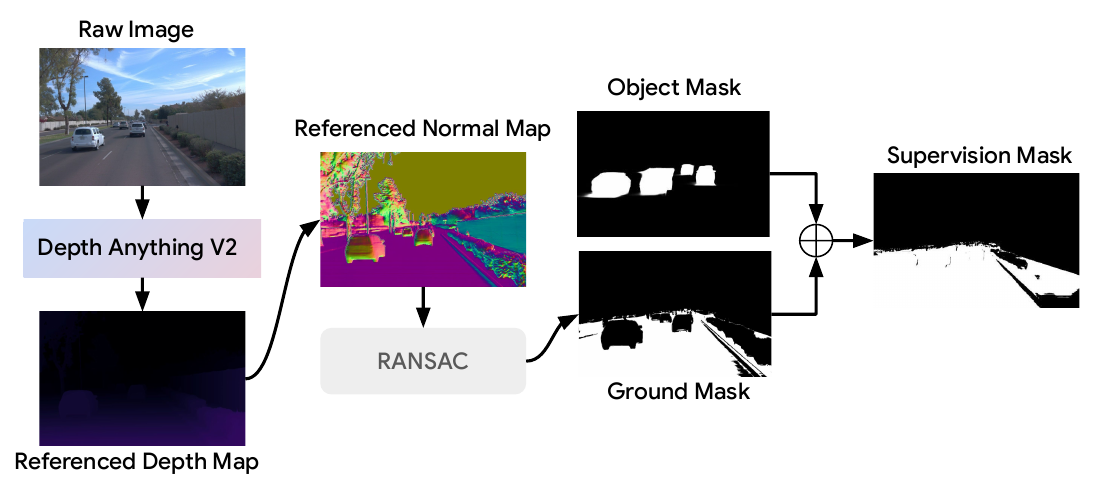}
    \caption{Construction of the supervision mask for geometry-consistent reconstruction. The raw image is processed by a depth foundation model to produce a reference depth map, from which a normal map is computed. The RANSAC-derived ground mask is fused with the preprocessed object mask to obtain the final supervision mask.}
    \label{fig:mask}
\end{figure}

\textbf{Normal and Geometry Constraints.}
To stabilize optimization in physically meaningful regions, we construct a supervision mask $\mathbf{M}$ by combining foreground objects and planar ground regions. Detailed construction is illustrated in Fig.~\ref{fig:mask}. Given the rendered normals $\mathbf{N}\in\mathbb{R}^{3\times H\times W}$ and the depth-derived reference normals $\mathbf{N}^{\ast}$, the normal constraint is defined via an angular loss
\begin{equation}
\mathcal{L}_{\text{normal}}
=
\frac{1}{|\mathbf{M}|}
\sum_{\mathbf{x}}
\mathbf{M}(\mathbf{x})
\left(1-
\left\langle
\mathbf{N}(\mathbf{x}),
\mathbf{N}^{\ast}(\mathbf{x})
\right\rangle
\right)
\end{equation}
where $\mathbf{N}(\mathbf{x})$ denotes the normal at pixel $\mathbf{x}$, and $\langle\cdot,\cdot\rangle$ represents the cosine similarity. This term enforces orientation consistency between rendered and geometry-induced normals.

To further encourage local surface coherence, we introduce a geometry consistency loss inspired by \cite{xie2025vid2sim}. Let $D\in\mathbb{R}^{H\times W}$ be the rendered depth map. We compute its Sobel gradient magnitude and normalize it to obtain a smoothness weight $\mathbf{W}_g\in[0,1]^{H\times W}$. We then compare each predicted normal with the locally averaged normal $\bar{\mathbf{N}}^{\ast}$ obtained via a $3\times3$ mean filter:
\[
\mathcal{L}_{\text{geo}}
=
\frac{\sum_{\mathbf{x}}
\mathbf{W}_g(\mathbf{x})
\left(1-
\left\langle
\mathbf{N}(\mathbf{x}),
\bar{\mathbf{N}}^{\ast}(\mathbf{x})
\right\rangle
\right)}
{\sum_{\mathbf{x}}\mathbf{W}_g(\mathbf{x})}
\]

By restricting supervision to mask $\mathbf{M}$ and jointly enforcing normal alignment and local geometric smoothness, the proposed constraints significantly improve surface quality on vehicles and road regions.

\subsection{LLM-Driven Corner Case Analysis and Synthesis}
\label{subsec:llm}
\begin{figure}[t]
    \centering
    \includegraphics[width=1.0\columnwidth]{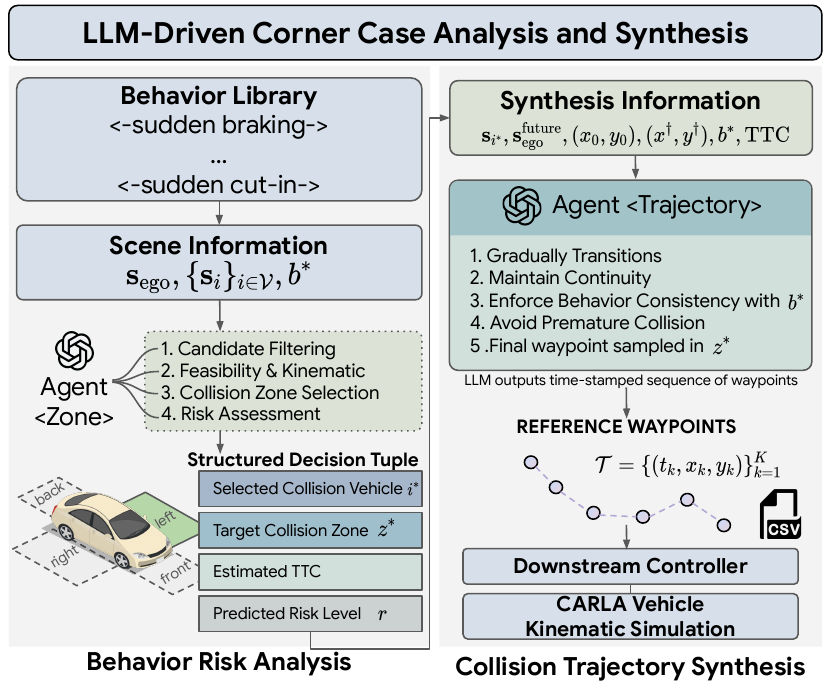}
    \caption{Overview of the LLM-driven corner-case generation pipeline. The system models ego-centric collision zones, performs risk analysis to select an adversarial vehicle and target zone, and synthesizes a waypoint-based collision-oriented trajectory that is later executed by CARLA for physically feasible simulation.}
    \label{fig:llm}
\end{figure}

To automatically construct safety-critical scenarios beyond simple trajectory perturbation, we develop a multi-agent LLM-driven corner case generation framework (as shown in Fig. \ref{fig:llm}) that explicitly targets collision-prone regions around the ego vehicle, enabling more controllable and spatially grounded safety-critical scenario synthesis. Given multi-vehicle tracking data and ego poses, we follow a chain-of-thought prompting paradigm \cite{wei2022chain} and design a pipeline consisting of the following stages:

\textbf{Collision Zone Modeling.}
We first define four rectangular collision zones tightly surrounding the ego vehicle in the ego-centric coordinate frame:
\begin{equation}
\mathcal{Z}=\{\text{front},\text{back},\text{left},\text{right}\}
\end{equation}
Each zone is parameterized by spatial bounds determined by the ego vehicle length $L_e$ and width $W_e$. These zones serve as explicit spatial targets for risk maximization and provide structured guidance to the language model.

\textbf{Risk Analysis and Collision Zone Selection.}
Given a user-specified dangerous behavior $b^\ast$ selected from a predefined behavior library, we employ a language model to perform scene-aware reasoning and identify both the most plausible adversarial vehicle and the most threatening collision region.

Let the ego vehicle be located at the origin of the ego-centric coordinate frame, and let $\mathcal{V}=\{1,\dots,N\}$ denote the set of observed background vehicles in the scene. For each vehicle $i\in\mathcal{V}$, we construct its recent motion state
\begin{equation}
\mathbf{s}_i=\{(x_t,y_t,v_t,\theta_t)\}_{t=1}^{T}
\end{equation}
where $(x_t,y_t)$ denotes the position in the ego-centric frame, $v_t$ is the speed, and $\theta_t$ is the heading over the past $T$ frames. The ego vehicle state is constructed in the same form and denoted by $\mathbf{s}_{\text{ego}}$.

The LLM agent then performs behavior-conditioned risk reasoning over all candidate vehicles. Formally, the decision process can be written as
\begin{equation}
(i^\ast,\; z^\ast,\; \mathrm{TTC},\; r)
=
\mathrm{Agent_{zone}}\!\left(
\mathbf{s}_{\text{ego}},
\{\mathbf{s}_i\}_{i\in\mathcal{V}},
b^\ast
\right)
\end{equation}
where $i^\ast\in\mathcal{V}$ is the selected adversarial vehicle, $z^\ast\in\mathcal{Z}$ is the predicted target collision zone chosen from the predefined zone set $\mathcal{Z}$ around the ego vehicle, $\mathrm{TTC}$ denotes the estimated time-to-collision in seconds assuming the dangerous behavior occurs, and $r\in\{\text{Low},\text{Medium},\text{High},\text{Critical}\}$ represents the risk level. In addition, the behavior $b^\ast$ can be either user-specified or automatically inferred by the LLM agent. Conditioning on a given behavior hypothesis narrows the reasoning space and improves generation controllability.

\textbf{Collision-Oriented Trajectory Synthesis.}
Conditioned on the selected adversarial vehicle and target zone $(i^\ast, z^\ast)$, we generate a reference trajectory using the LLM. Instead of predicting dynamically feasible control signals, which lies outside the core competency of language models, we delegate low-level feasibility to a downstream controller and allow the LLM to focus on high-level waypoint generation.

Let $(x_0,y_0)$ denote the current position of vehicle $i^\ast$ in the ego-centric frame, and let $\mathbf{s}_{i^\ast}$ and $\mathbf{s}_{\text{ego}}^{\text{future}}$ denote the historical trajectory of the adversarial vehicle and the predicted future ego trajectory, respectively. A collision target point $(x^{\dagger},y^{\dagger})$ is sampled within the selected zone $z^\ast$. The waypoint generation process is formulated as
\begin{equation}
\mathcal{T}
=
\mathrm{Agent_{traj}}\!\left(
\mathbf{s}_{i^\ast},
\mathbf{s}_{\text{ego}}^{\text{future}},
(x_0,y_0),
(x^{\dagger},y^{\dagger}),
b^\ast,
\mathrm{TTC}
\right)
\end{equation}
where
\begin{equation}
\mathcal{T}=\{(t_k,x_k,y_k)\}_{k=1}^{K}
\end{equation}
is the generated time-stamped waypoint sequence over the prediction horizon.
The final waypoint is constrained to coincide with the sampled collision target. The resulting waypoint sequence can be directly interfaced with CARLA for vehicle kinematic simulation.

\subsection{Real-Sim-Real: Bridging CARLA and Gaussian Splatting}
\label{subsec:carla}
To obtain physically consistent corner-case visualizations, we adopt a Real-Sim-Real pipeline that bridges the LLM-generated trajectory, CARLA vehicle dynamics, and Gaussian-based rendering. 

\textbf{CARLA-based Kinematic Simulation.}
Given the LLM-generated waypoint sequence $\mathcal{T}$ from the last step, we first import the trajectory into CARLA as a reference path. Then we design a PID controller that tracks the reference waypoints in continuous time for the selected vehicle $i^\ast$. Specifically, the control command at time $t$ is computed as
\begin{equation}
\mathbf{u}_t = \mathrm{PID}\big(\mathbf{p}_t,\;\mathbf{p}_t^{\text{ref}}\big)
\end{equation}
where $\mathbf{p}_t$ and $\mathbf{p}_t^{\text{ref}}$ denote the current and reference vehicle poses, respectively. The simulator then evolves the vehicle state according to its internal kinematic model
\begin{equation}
\mathbf{x}_{t+1} = f_{\text{CARLA}}(\mathbf{x}_t,\mathbf{u}_t)
\end{equation}
producing a physically feasible trajectory
\begin{equation}
\tilde{\mathcal{T}}=\{(t_k,\tilde{x}_k,\tilde{y}_k,\tilde{\theta}_k)\}_{k=1}^{K}
\end{equation}
This step ensures dynamic realism and smooth vehicle motion. In practice, we employ the built-in \texttt{VehiclePIDController} to track the reference path. 

\textbf{Back-Projection to Gaussian Space.}
After simulation, the CARLA-generated trajectory $\tilde{\mathcal{T}}$ is mapped back to the Gaussian scene representation. For each timestamp, we update the pose of the corresponding Gaussian actor using the simulated position $(\tilde{x}_k,\tilde{y}_k)$ and heading $\tilde{\theta}_k$. The modified Gaussian scene is finally rendered using the differentiable Gaussian renderer.

\begin{figure}[t]
    \centering
    \includegraphics[width=1.0\columnwidth]{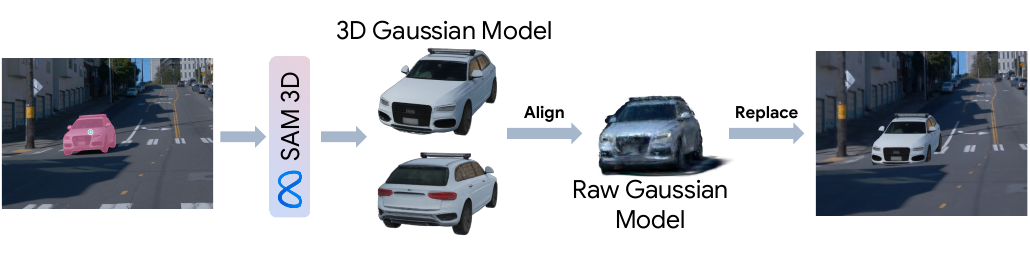}
    \caption{Vehicle replacement pipeline. We first select frames where the target vehicle is close to the camera, and reconstruct a 3D Gaussian model using SAM 3D. The reconstructed asset is aligned with the original vehicle geometry and replaces it.}
    \label{fig:sam3d}
\end{figure}

\begin{table*}[t]
\centering
\caption{Ablation study on rendering quality and geometry consistency.}
\label{tab:render_quality}
\setlength{\tabcolsep}{6pt}
\renewcommand{\arraystretch}{1.12}
\begin{tabular}{lccccc}
\toprule
\textbf{Components} 
& \textbf{PSNR $\uparrow$} 
& \textbf{SSIM $\uparrow$} 
& \textbf{LPIPS $\downarrow$} 
& \textbf{Normal L1 MAE $\downarrow$} 
& \textbf{Normal Cos MAE $\downarrow$} \\
\midrule
Baseline
& 29.703 
& 0.907 
& 0.219 
& 0.682 
& 0.984 \\

+$\mathcal{L}_{\mathrm{flat}}$ 
& \textbf{29.828} 
& \textbf{0.908} 
& \textbf{0.215} 
& 0.677 
& 0.957 \\

+$\mathcal{L}_{\mathrm{flat}} + \mathcal{L}_{\mathrm{normal}}$ 
& 29.521 
& 0.906 
& 0.217 
& 0.361 
& 0.482 \\

+$\mathcal{L}_{\mathrm{flat}} + \mathcal{L}_{\mathrm{normal}} + \mathcal{L}_{\mathrm{geo}}$ 
& 29.027 
& 0.900 
& 0.223 
& \textbf{0.350} 
& \textbf{0.480} \\
\bottomrule
\end{tabular}
\end{table*}

\begin{figure*}[t]
    \centering
    \includegraphics[width=1.0\textwidth]{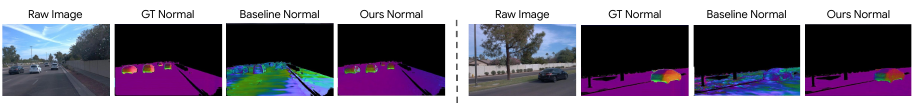}
    \caption{Surface normal renderings. Our geometry-consistency constraints produce finer surface details and fewer artifacts than the baseline.}
    \label{fig:normal}
\end{figure*}

\textbf{Vehicle Replacement with 3D Foundation Models.}
Since the Gaussian model is trained from limited camera views, rendering unseen viewpoints may introduce tearing artifacts, especially for near-field vehicles with insufficient observations (see Fig.~\ref{fig:corner}). To mitigate this, we replace such vehicles with high-quality assets reconstructed by SAM-3D~\cite{chen2025sam}, as shown in Fig. \ref{fig:sam3d}.

Specifically, we use SAM-3D~\cite{chen2025sam} to reconstruct a 3D Gaussian representation from close-range frames containing the target vehicle. To align the reconstructed asset with the original object, we denote the original and replacement point sets as $\mathcal{P}_o$ and $\mathcal{P}_r$. Both point clouds are first centered. Their yaw orientations $\theta_o$ and $\theta_r$ are estimated from the minimum-area bounding boxes of their XY projections, producing a relative yaw difference $\Delta\theta=\theta_o-\theta_r$. To account for symmetry ambiguity, we evaluate a small set of candidate rotations $\Theta=\{\Delta\theta,\Delta\theta+\pi,\Delta\theta+\frac{\pi}{2},\Delta\theta-\frac{\pi}{2}\}$. The object scale is matched by comparing the bounding-box extents $\mathbf{e}_o$ and $\mathbf{e}_r$ and computing the scale factor $\mathbf{s}=\mathbf{e}_o/\mathbf{e}_r$. The final alignment is selected by minimizing
\begin{equation}
E(\theta) =
\|\mathbf{e}(\hat{\mathcal{P}}_r)-\mathbf{e}(\mathcal{P}_o)\|_2
+
\lambda\, d_{\text{Chamfer}}(\hat{\mathcal{P}}_r,\mathcal{P}_o),
\theta\in\Theta
\end{equation}
where $\hat{\mathcal{P}}_r$ is the transformed replacement point set and $d_{\text{Chamfer}}(\cdot)$ denotes the bidirectional nearest-neighbor distance between the two point clouds. This procedure enables the reconstructed vehicle to replace the original object while preserving geometric consistency and visual realism.

\section{EXPERIMENTS}
\subsection{Experimental Setup}
We adopt Street Gaussians \cite{yan2024street} as our baseline, whose object-background modeling strategy provides a strong foundation for instance-level vehicle editing in our setting. We conduct experiments on the Waymo Open Dataset~\cite{sun2020scalability}, selecting 8 sequences to synthesize 85 corner-case scenarios. For Gaussian training, we follow the training protocol of \cite{yan2024street} for most settings, with a total of 50,000 iterations. To reduce training time, we only use three cameras, [front, front left, front right], and incorporate the three proposed geometric consistency loss terms, each weighted by 0.1. For LLM-based corner-case reference trajectory generation, we use the ChatGPT-5.2 API. Both Gaussian rendering and LLM trajectory planning are executed at 24 FPS. For vehicle physical simulation, we use CARLA 0.9.16, with the longitudinal and lateral PID gains set to $k_p=2.0$, $k_d=0.2$, $k_i=0.0$, and a simulation step size of $\Delta t=0.2 \text{s}$.

\subsection{Quantitative Evaluation}
\textbf{Corner-case Controllability and Success.}
We evaluate the controllability and effectiveness of generated corner cases against rule-based and random trajectory baselines. The rule-based policy triggers predefined maneuvers from the initial relative vehicle positions, while the random policy uses stochastic perturbations without semantic reasoning. All methods use identical initial conditions and horizons. Since few existing methods closely match our task design and in-scene corner-case generation is inherently stochastic, direct comparison with other state-of-the-art methods is difficult.

Table~\ref{tab:cc_success} reports three metrics: \emph{Zone Hit} (whether the trajectory reaches the target collision zone), \emph{Success} (whether a safety-critical interaction occurs with $\text{TTC}<1$\,s), and \emph{MinTTC} (the minimum time-to-collision in the scenario). Our method achieves the best controllability, with a Zone Hit rate of $0.438$, exceeding the rule-based ($0.138$) and random ($0.063$) baselines. It also produces the most critical interactions, achieving the highest Success rate ($0.925$) and lowest MinTTC ($0.472$ s), demonstrating the effectiveness of the proposed multi-agent LLM framework.

However, LLM instability can still limit planning reliability. As indicated by the zone hit rate and variance, the LLM may fail to reach the target zone under multiple constraints, even with explicit prompts, as shown in Case 3 of Fig.~\ref{fig:llm_traj}. Manual inspection shows that 29.4\% of LLM-generated outputs are valid.

\begin{table}[t]
\centering
\caption{Corner-case controllability and success statistics. Results are reported as mean $\pm$ std.}
\label{tab:cc_success}
\setlength{\tabcolsep}{5pt}
\renewcommand{\arraystretch}{1.12}
\resizebox{\columnwidth}{!}{
\begin{tabular}{lccc}
\toprule
\textbf{Method} 
& \textbf{Zone Hit $\uparrow$} 
& \textbf{Success $\uparrow$} 
& \textbf{MinTTC (s) $\downarrow$} \\
\midrule
Rule-based 
& $0.138 \pm 0.347$ 
& $0.613 \pm 0.490$ 
& $1.483 \pm 1.794$ \\
Random 
& $0.063 \pm 0.244$ 
& $0.738 \pm 0.443$ 
& $1.095 \pm 1.560$ \\
\textbf{Ours (Multi-agent LLM)} 
& $\mathbf{0.438 \pm 0.499}$ 
& $\mathbf{0.925 \pm 0.265}$ 
& $\mathbf{0.472 \pm 1.215}$ \\
\bottomrule
\end{tabular}}
\end{table}

\begin{table}[t]
\centering
\caption{Feasibility and ride-comfort metrics.}
\label{tab:comfort}
\setlength{\tabcolsep}{4pt}
\renewcommand{\arraystretch}{1.12}
\resizebox{\columnwidth}{!}{
\begin{tabular}{lccc}
\toprule
\textbf{Method} 
& $\boldsymbol{|a^{\mathrm{lat}}|_{95}}$ $\mathbf{(m/s^2)}$ $\downarrow$ 
& $\boldsymbol{|\dot{\kappa}|_{95}}$ $\mathbf{(1/(m\cdot s))}$ $\downarrow$ 
& \textbf{ComfortViol (\%) $\downarrow$} \\
\midrule
LLM-only 
& $2.66 \pm 5.17$ 
& $1.02 \pm 4.05$ 
& $9.62 \pm 19.15$ \\
\textbf{Ours (LLM+CARLA)} 
& $\mathbf{2.02 \pm 3.35}$ 
& $\mathbf{0.43 \pm 0.97}$ 
& $\mathbf{9.39 \pm 21.45}$ \\
\bottomrule
\end{tabular}}
\end{table}

\textbf{Trajectory Feasibility and Ride Comfort.}
We further evaluate the physical feasibility and ride comfort of generated trajectories. 
Three commonly used kinematic indicators are computed from time-resampled trajectories: 
the $95$-percentile lateral acceleration $|a^{lat}|_{95}$, the $95$-percentile curvature-rate $|\dot{\kappa}|_{95}$ measuring steering smoothness, and the percentage of comfort violations (ComfortViol) where $|a^{lat}|>3$ m/s$^2$.
Table~\ref{tab:comfort} compares trajectories generated directly by the LLM and those executed through CARLA. By delegating motion execution to CARLA, the proposed pipeline improves all metrics, reducing $|a^{lat}|_{95}$ ($2.66\rightarrow2.02$) and $|\dot{\kappa}|_{95}$ ($1.02\rightarrow0.43$), and slightly decreasing comfort violations. This indicates that physics-based control effectively regularizes LLM-generated trajectories and yields smoother and more feasible vehicle behaviors.

\subsection{Qualitative Results}
\textbf{Reconstruction Evaluation.}
Fig.~\ref{fig:normal} shows that the geometric consistency constraint produces smoother surfaces and normals closer to the ground truth. Table~\ref{tab:render_quality} further reports PSNR, SSIM, LPIPS, Normal L1 MAE, and Normal Cos MAE under different constraints. Compared with the baseline, the proposed constraints substantially reduce normal errors while causing only negligible changes in rendering quality, demonstrating improved geometric consistency without sacrificing photometric fidelity.

\begin{figure}[t]
    \centering
    \includegraphics[width=1.0\columnwidth]{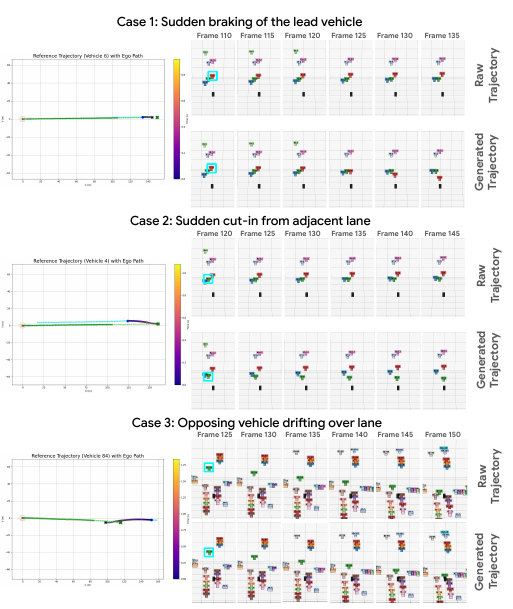}
    \caption{Visualization of LLM-driven corner-case trajectory generation. Left: historical and predicted trajectories of the ego vehicle and the selected vehicle. Green and light-green points denote the ego vehicle’s past and future trajectories, while blue and color-gradient points denote the selected vehicle’s past and LLM-generated future trajectories. The green cross indicates the collision target point. Right: corresponding 3D scene snapshots over time, where the red box highlights the selected vehicle. The generated trajectories exhibit representative corner-case behaviors (sudden brake, cut-in, and lane drift), providing reference motion cues for downstream CARLA-based kinematic simulation.}
    \label{fig:llm_traj}
\end{figure}

\begin{figure}[t]
    \centering
    \includegraphics[width=1.0\columnwidth]{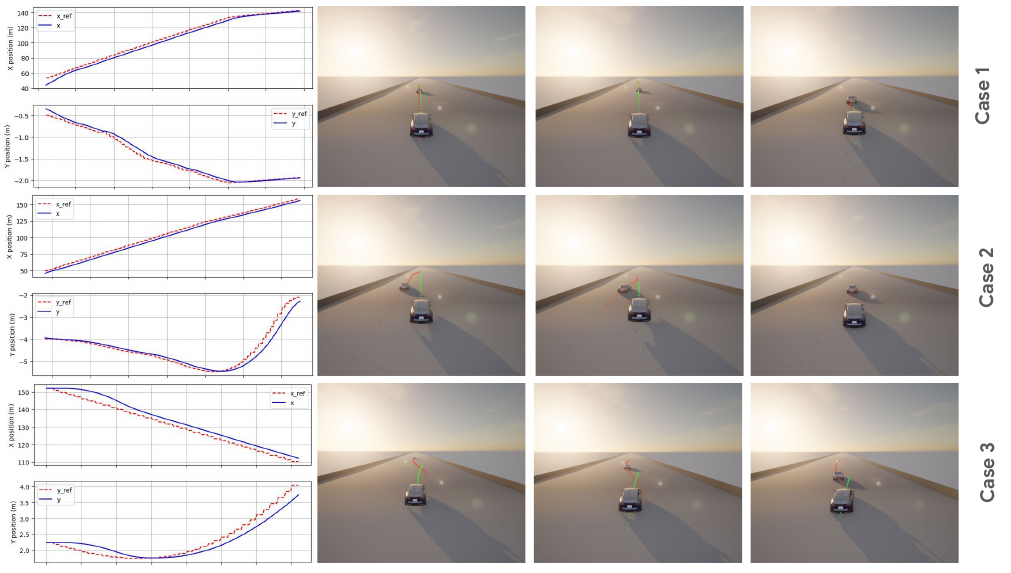}
    \caption{CARLA-based realization of LLM-generated trajectories. Left: PID tracking performance along x (front) and y (left). Right: CARLA simulation. The ego trajectory is shown in green, and the selected vehicle follows the red reference points under continuous PID control.}
    \label{fig:carla}
\end{figure}

\textbf{LLM Corner Case Generation.}
\begin{figure}[t]
    \centering
    \includegraphics[width=1.0\columnwidth]{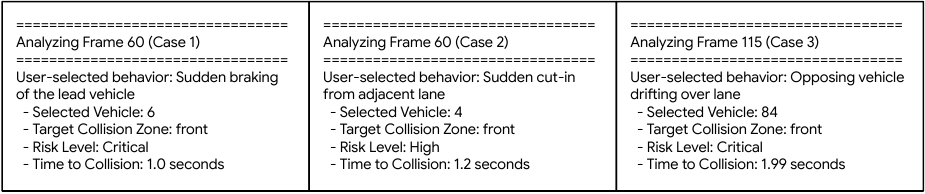}
    \caption{Behavior risk analysis results produced by the LLM agent. Given the starting analysis frame and scene context, the Agent $<$Zone$>$ outputs the target vehicle, collision zone, risk level, and TTC.}
    \label{fig:llm_out}
\end{figure}
We manually define three corner-case types, sudden brake, cut-in, and lane drift, and provide them to the LLM to analyze the scene and generate reference trajectories. Fig.~\ref{fig:llm_out} shows the outputs of Agent $<$Zone$>$, while Fig.~\ref{fig:llm_traj} visualizes the past and predicted trajectories of the ego vehicle and the selected vehicle produced by Agent $<$Trajectory$>$.

\textit{Case 1.} Starting from frame 60, Agent $<$Zone$>$ identifies vehicle 6 ahead of the ego vehicle, selects the collision zone as “front”, and estimates TTC as 1.0\,s. Agent $<$Trajectory$>$ then generates a braking trajectory toward the collision target point, producing a near rear-end interaction compared with the safe headway observed in the original trajectory. \textit{Case 2 and Case 3.} The remaining cases illustrate cut-in and lane-drift behaviors: vehicle 4 performs a rapid lane change under short headway, and oncoming vehicle 84 crosses the lane toward the ego vehicle. These examples show that the LLM agents can identify risky participants and generate behavior-consistent trajectories for subsequent simulation in CARLA.

\begin{figure*}[t]
    \centering
    \includegraphics[width=1.0\textwidth]{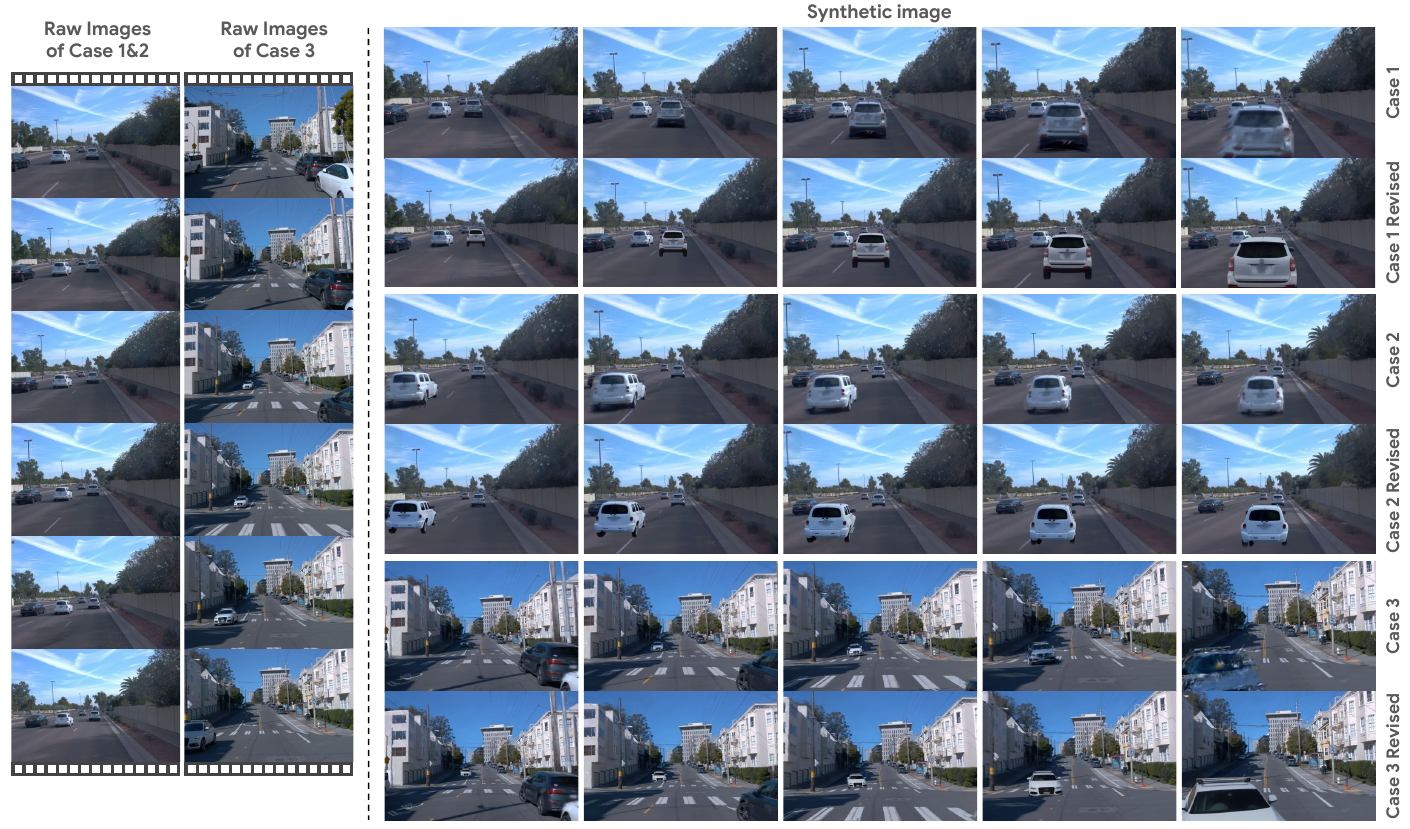}
    \caption{Final Gaussian renderings of the three synthesized corner cases. Left: raw input frames. Right: edited renderings after LLM-driven reasoning and CARLA-based physical simulation. For one case, we show both the vanilla version and the result after replacing the vehicle with a 3D asset reconstructed by SAM-3D. The results demonstrate that the proposed pipeline can generate realistic and semantically consistent corner-case scenarios, while model replacement further improves the geometric and visual consistency of the rendered vehicles.}
    \label{fig:corner}
\end{figure*}

\textbf{CARLA Simulation.}
We feed the LLM-generated reference waypoints into CARLA and enforce physically plausible vehicle motion using a PID controller. Specifically, we implement a frame-by-frame tracking controller that drives the selected vehicle to follow the reference trajectory. Fig.~\ref{fig:carla} illustrates this process. The left plot shows PID tracking performance, while the right panel presents the corresponding CARLA snapshots. The results demonstrate that CARLA can reproduce the LLM-generated trajectory while providing physically realistic motion through its physics engine.

Due to the inherent limitations of PID control, a small tracking gap exists between the executed and reference trajectories. This discrepancy does not affect the validity of the simulation, since the goal of our framework is to synthesize previously unseen corner cases from real-world data. To further reduce tracking error, PID tracking is initialized from frame 0 rather than the LLM input start frame, allowing the controller to stabilize before the corner-case generation interval.

\textbf{Sim-to-Real: From CARLA to Gaussian Splatting.}
After obtaining the simulated trajectories and poses from CARLA, we re-project them into the Gaussian representation. Owing to the independent vehicle modeling, the simulated states can be seamlessly injected by replacing the selected vehicle’s data. Fig.~\ref{fig:corner} shows the final Gaussian renderings for all three cases, demonstrating faithful alignment with both the LLM reasoning outcomes and the CARLA-based physically consistent motions, resulting in high-fidelity and semantically consistent corner-case scenarios. The replacement of vehicles with SAM-3D reconstructed models further improves the geometric and visual consistency of the rendered vehicles.

\section{Computational Cost Analysis}
The CARLA-GS pipeline consists of several stages. To evaluate its practical feasibility, we summarize the computational cost of each component here. Prior information, such as depth, normal, and mask maps, is prepared in advance and excluded from the online deployment time. The main bottleneck is 3DGS reconstruction, which takes about 2 hours for a 100-frame, 3-camera scene on a single RTX 4090 GPU. LLM calls require about 40--50 seconds for scene reasoning and 10 seconds for trajectory generation. CARLA simulation time depends on the synthesized trajectory length and vehicle kinematic settings, but a single PID-controlled rollout usually takes less than 60 seconds. Backprojection and SAM-3D replacement are excluded since 3DGS rendering is fast and the 3D asset library is pre-built.

Overall, 3DGS training dominates the cost but can be accelerated by recent fast 3DGS pipelines. Other stages are lightweight, and multiple corner cases can be synthesized per scene, making CARLA-GS safer and more time-efficient than real-world data collection.

\section{CONCLUSIONS}
This paper presented CARLA-GS, a modular framework for safety-critical corner-case synthesis that tightly integrates Gaussian-based visual reconstruction, LLM-driven semantic reasoning, and physics-consistent execution in CARLA. By explicitly decoupling representation, reasoning, and dynamics while preserving cross-module consistency, the proposed pipeline enables controllable and physically valid generation of rare but safety-critical traffic interactions from real-world data. Experiments on the Waymo Open Dataset demonstrate that CARLA-GS can reliably synthesize diverse corner cases, while maintaining photorealistic rendering quality and kinematic feasibility. The results highlight the effectiveness of leveraging complementary strengths of neural rendering, large language models, and mature vehicle simulators within a unified framework.

Future work will explore stronger view-consistency constraints under sparse camera settings, closed-loop verification of LLM reasoning, large-scale automatic mining of high-risk scenarios from long-horizon traffic data, and evaluation on downstream tasks.

% \section*{APPENDIX}
% \section*{ACKNOWLEDGMENT}

\bibliographystyle{IEEEtran}
\bibliography{ref}

\end{document}